# An Improved End-to-End Multi-Target Tracking Method Based on Transformer Self-Attention

Yong Hong [1,3], Deren Li [1, *], Shupei Luo [2], Xin Chen [2], Yi Yang [2], Mi Wang [1]

[1] State Key Laboratory of Information Engineering in Surveying, Mapping and Remote Sensing,Wuhan University, Wuhan 430079, China; 2019106190017@whu.edu.cn (Y.H.); wangmi@whu.edu.cn (M.W.);

[2] Wuhan Optics Valley Information Technology Co., Ltd., Wuhan, Hubei, 430068, China; luoshup@51bsi.com (S.L.); chenxinx@51bsi.com (X.C.);yangyi@51bsi.com (Y.Y.);

[3] Mobile Broadcasting and Information Service Industry Innovation Research Institute (Wuhan) Co., Ltd, Wuhan, Hubei, 430068, China;

* Correspondence: drli@whu.edu.cn Tel: (+86-13907144816)(D.L.);

**Abstract**: This study proposes an improved end-to-end multi-target tracking algorithm that adapts to multi-view multi-scale scenes based on the self-attentive mechanism of the transformer's encoder-decoder structure. A multi-dimensional feature extraction backbone network is combined with a self-built semantic raster map, which is stored in the encoder for correlation and generates target position encoding and multi-dimensional feature vectors. The decoder incorporates four methods: spatial clustering and semantic filtering of multi-view targets, dynamic matching of multi-dimensional features, space-time logic-based multi-target tracking, and space-time convergence network (STCN)-based parameter passing. Through the fusion of multiple decoding methods, muti-camera targets are tracked in three dimensions: temporal logic, spatial logic, and feature matching. For the MOT17 dataset, this study's method significantly outperforms the current state-of-the-art method MiniTrackV2 [49] by 2.2% to 0.836 on Multiple Object Tracking Accuracy(MOTA) metric. Furthermore, this study proposes a retrospective mechanism for the first time, and adopts a reverse-order processing method to optimise the historical mislabeled targets for improving the Identification F1-score(IDF1). For the self-built dataset OVIT-MOT01, the IDF1 improves from 0.948 to 0.967, and the Multi-camera Tracking Accuracy(MCTA) improves from 0.878 to 0.909, which significantly improves the continuous tracking accuracy and scene adaptation. This research method introduces a new attentional tracking paradigm which is able to achieve state-of-the-art performance on multi-target tracking (MOT17 and

OVIT-MOT01) tasks.

# 1 Introduction

## 1.1 Research Background

Vision-based multi-target multi-camera tracking (MTMCT) algorithms for pedestrians have become an area of interest for many researchers, but numerous problems still exist. Several examples include low target re-identification rates across cameras in multi-view and multi-scale scenarios[1], one target with different texture features at different views, and a single-dimensional feature extraction and matching algorithm that cannot be adapted to this scenario[2]. To solve the problems of MTMCT , there are usually two paradigms: "tracking-by-detection" (TBD) and "tracking-by-attention" (TBA) .The TBD paradigm-based algorithm[3-7] has a complex structure, is difficult to deploy, and is inefficient in multi-camera scenarios.And the Existing methods of TBA paradigm-based algorithm[9-10] is relatively simple and cannot be applied to complex scenes.An end-to-end target tracking algorithm that can adapt to multi-view and multi-scale scenarios can help solve these problems.

## 1.2 Related work

Figure 1 shows the target-tracking algorithm structure of the TBD paradigm. Most traditional multi-target tracking methods folow the paradigm[3-7]. The overall logic of the TBD paradigm is heuristic and simple, resulting in difficulties in modeling the spatio-temporal complexities of the target and low robustness in the multi-view and multi-scale scenarios across cameras.

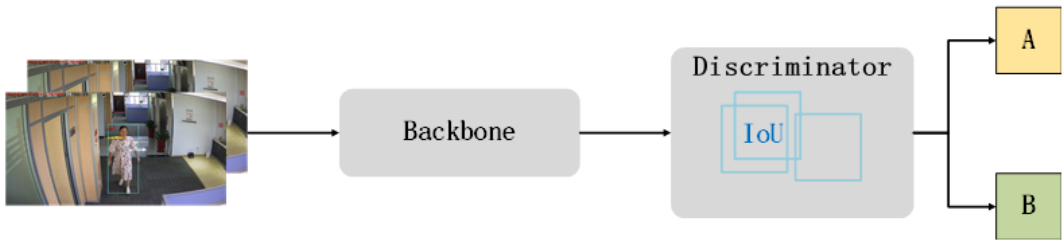

(a) Spatial similarity-based association methods

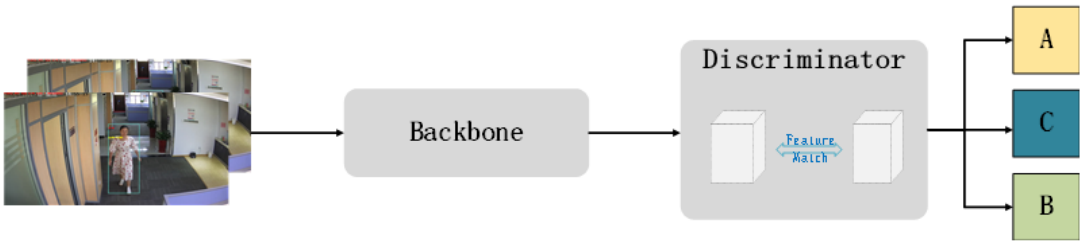

(b) Association methods based on similarity of appearance

**Figure 1** algorithm structure of the TBD paradigm

The rise of transformer-based multi-target tracking networks[8] has ushered in a new paradigm called "tracking-by-attention" (TBA)(see Figure 2). Tim M. et al.'s TrackFormer[9] network achieved seamless data association between frames by implementing both position masking and object identity inference through an encoder-decoder self-attention mechanism. Zeng[10] developed a temporal aggregation network for passing temporal correlation query information to aid continuous tracking. The self-attention mechanism of the transformer-based multi-target tracking algorithm is extremely informative, but its decoder construction is relatively simple and cannot be applied to complex scenes across cameras.

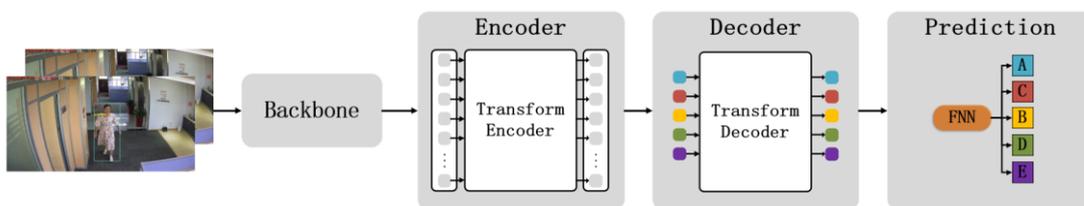

**Figure 2** algorithm structure of the TBA paradigm

In the field of MTMCT, traditional algorithms are discussed in two scenarios: overlapping field of view(FOVs) and non-overlapping FOVs. For example, in the case of overlapping fields of view, Berclaz et al.[11] solved the association of targets based on the shortest k-path algorithm, and Hu et al.[12] matched cross-camera trajectories based on pairwise geometric constraints. In the case of non-overlapping fields of view, Cai et al.[13] matched different targets based on appearance similarity comparison, while Ristani et al. [14] extracted appearance and motion features based on Convolutional Neural Network(CNN) networks to complete target Re-identification(ReID).

At the same time, Markov random fields and conditional random fields have been frequently used for target association. For example, Chen et al. [15] used Markov chains and Monte Carlo sampling to obtain the space-time location relationship between cameras and image features for inter-camera association. Chen et al. [16] used Markov random fields to construct an equalized graph model. Chen and Bhanu[17] used conditional random fields to correlate tracklets generated by a single camera. Lee et al. [18] used inter camera linking model to match temporal, regional, and fusion features. The above methods are limited to solving target re-identification in specific scenes but cannot be applied in complex scenes with multiple views and scales. Thus, generalization needs to be improved.

## 1.3  Problems

The problems of MTMCT in complex scenes are as follows.

**1.3.1 Accuracy enhancement for ReID under complex conditions**

Multiple targets in cross-camera situations, where the target is obscured by the scene or the target moves across the camera, can cause the network to lose continuous tracking of the target. When the target moves within the scene, the changes of position from far to near and the rotation can lead to changes in the scale and viewpoint of the target. All of the above situations will affect the ReID accuracy.

**1.3.2 Normalization of complex scene**

There are two scenarios for cross-camera target ReID: overlapping FOVs and non-overlapping FOVs, and the continuous tracking task is divided into three scenarios: "single camera target tracking", "cross-camera overlapping FOVs target tracking" and "cross-camera non-overlapping FOVs target tracking". Different scenarios have specific requirements for the algorithm mechanism, and traditional methods usually implement three different algorithms to solve the three scenarios separately[48], which is inefficient. Implementing a full-scene normalization algorithm can improve the overall efficiency and reduce deployment resource consumption.

## 2  Materials and methods

### 2.1 Materials

**2.1.1 Image data**

The MOT17[19] public dataset was used in this study, along with the self-built loop-tracking dataset OVIT-MOT01.

MOT17 is a standard dataset proposed in 2017 for measuring multi-target detection and tracking methods.

The self-built loop tracking dataset OVIT-MOT01 was constructed from video captured by five cameras, arranged in a zigzag office area, and calibrated for internal and external orientation. It contains 10,105 consecutive images and 8299 detection frames to evaluate the accuracy of cross-camera pedestrian re-identification and tracking.

**2.1.2 Raster semantic map data construction**

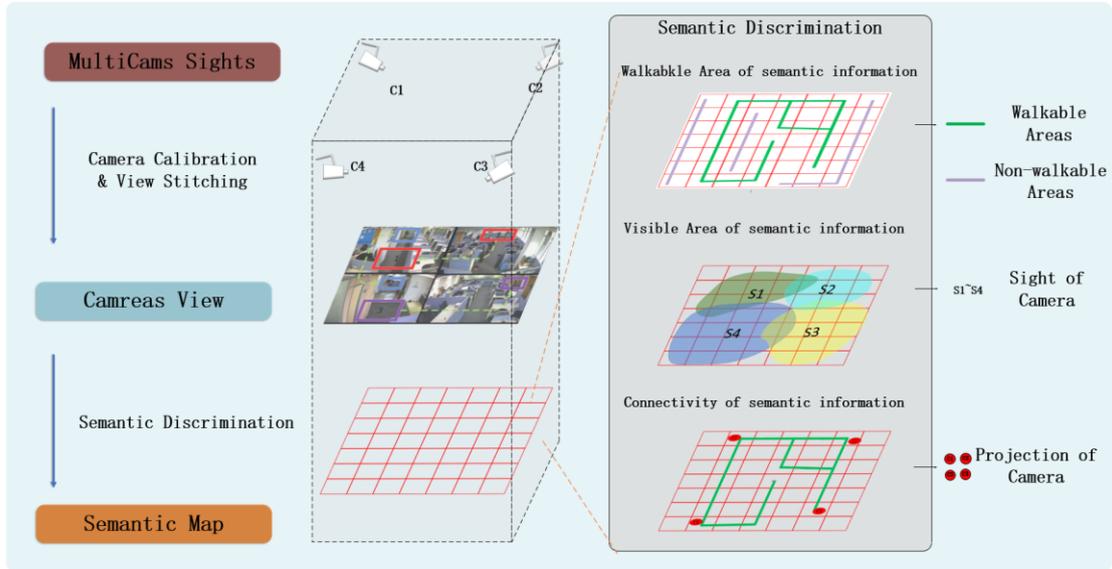

**Figure 3**  Semantic raster map construction

Figure 3 presents the general procedure for generating the semantic raster maps [21~23]. The main steps are as follows:

a) Joint calibration and area association stitching based on multiple view cameras in the scene were used to obtain a location association map in the corresponding pixel space.

b) The global image based on the camera pixel space resolution was gridded to obtain a raster vector base map. A pointer matrix $C_{ij}$ based on the raster with coordinates (i, j) was then constructed to represent the raster attributes.

c) Semantic information was sequentially generated on the co-visibility, walkability, and connectivity of the raster semantic map[24-26].

d) Semantic information on the co-visibility of raster maps based on camera calibration parameters, the projection of the camera field of view into object space S1 to S4, and information on the currently visible raster were recorded.

e) Semantic information on the walkability and connectivity of the raster map was projected onto the raster map based on the base map or motion trajectory. The walkability and connectivity information of the current raster was then recorded.

## 2.2 Methods

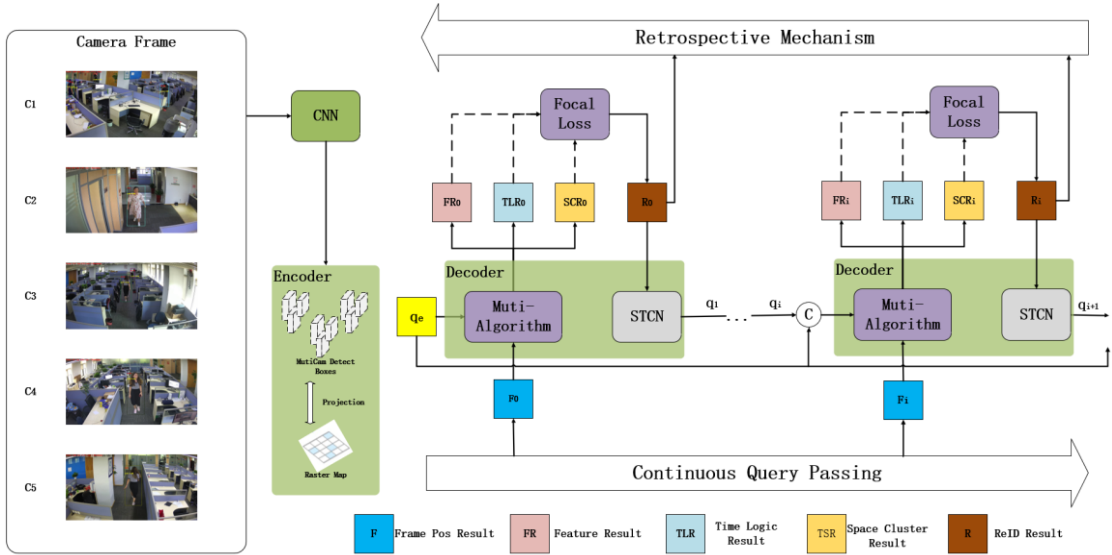

**Figure 4**  Overall algorithm structure

In this study, an end-to-end multi-objective tracking method was proposed based on the transformer's self-attentive improvement. The proposed approach continuously tracked pedestrians in complex situations (e.g., cross-camera, multi-view, and multi-scale) and constructed a raster semantic map to encode target locations. Cross-camera targets were continuously tracked based on three dimensions (i.e., temporal, spatial and logical), and a Space-Time Convergence Network (STCN) was constructed to transfer relevant feature parameters. A back-viewing mechanism was also added to optimize the historical target tracking results by inverting the order and normalizing the "overlapping field of view" and "non-overlapping field of view" scenarios in cross-camera tracking. The steps are as follows:

    a) For the multi-view camera detection, the corresponding detection frames and texture features were first obtained based on a multi-dimensional feature extraction CNN network and fed into the encoder.

    b) The encoder received the raster semantic map, which was constructed based on the target scene. Using the projection of the multi-dimensional feature detection frame and the raster semantic map from the multi-view detection results, the final detection frame result (Frame Pos Result) in the object space was

obtained and sent into the decoder.

c) The decoder received the frame detection result from the encoder and the a priori query of the previous frame. The decoder consisted of three parts: the spatial clustering and semantic filtering algorithm that generated the spatial clustering results, the multi-dimensional feature dynamic matching algorithm combined with the raster semantic map filter that produced the feature, and the space-time logic-based multi-visual target tracking algorithm that created the logic result. The results are subjected to cross-entropy loss to obtain the continuous tracking ReID result of the current frame, which is input to the STCN.

d) The STCN produced an a priori query for the next frame and cascaded it with the historical query. It then feeds it into the decoder and repeats (1)(2)(3) in the algorithm for the next frame.

e) When the overall tracking had been completed, the ReID was optimized by reviewing the overall results using inverted order processing and compensating for the confidence score in the historical results.

### 2.2.1 Backbone network construction for target detection based on multi-dimensional feature extraction

In the proposed method, YOLOV5[29] and ResNet50[27] are fused to construct a multi-dimensional feature extraction backbone network based on the pre-trained convolutional neural network YOLOV5. Pedestrian and head detection frames are extracted using regression to obtain candidate target locations, while multi-target texture features based on ResNet50 [28] are obtained to produce multi-dimensional feature vectors[28].

### 2.2.2 Construction of a transformer-based encoder

In constructing the transformer-based encoder, the raster vector is associated with the camera field of view based on pre-constructed raster semantic map data to obtain the position encoding of the target in a particular scene and fuse it with a multi-dimensional feature vector. The encoder outputs are the target position encoding and the multi-dimensional feature vector.

### 2.2.3 Construction of a transformer-based decoder

The decoder is divided into four steps: raster-based semantic map-assisted filtering algorithm, multi-dimensional feature dynamic matching, space-time logic-based multi-target tracking and STCN-based parameter passing. The overall structure is shown in Figure 5 below.

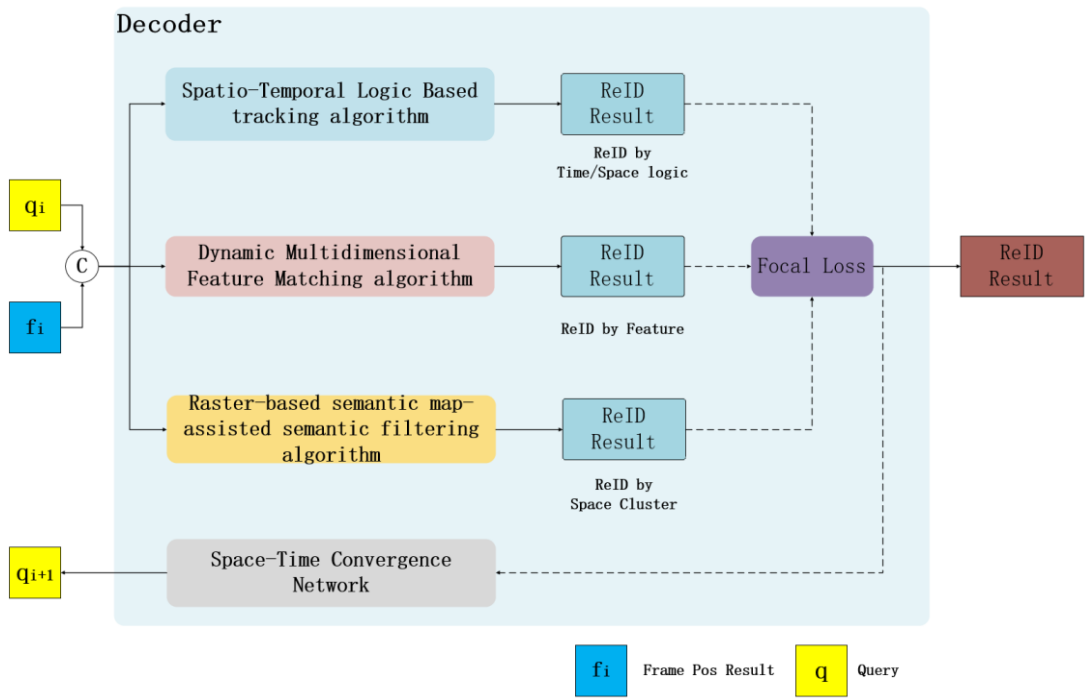

**Figure 5** Decoder structure

#### 2.2.3.1 Raster-based semantic map-assisted semantic filtering algorithm

A filtering algorithm was developed based on three types of semantic information (co-visibility, walkability and connectivity) for semantic raster map and used in the decoder.

**(1) Multi-visual target space clustering algorithm based on co-visuality**

The targets are projected onto a global raster semantic map based on the position encoding input. The targets in the overlapping field of view area are clustered using the co-visibility semantic information to complete coarse matching, as shown in Figure 6.

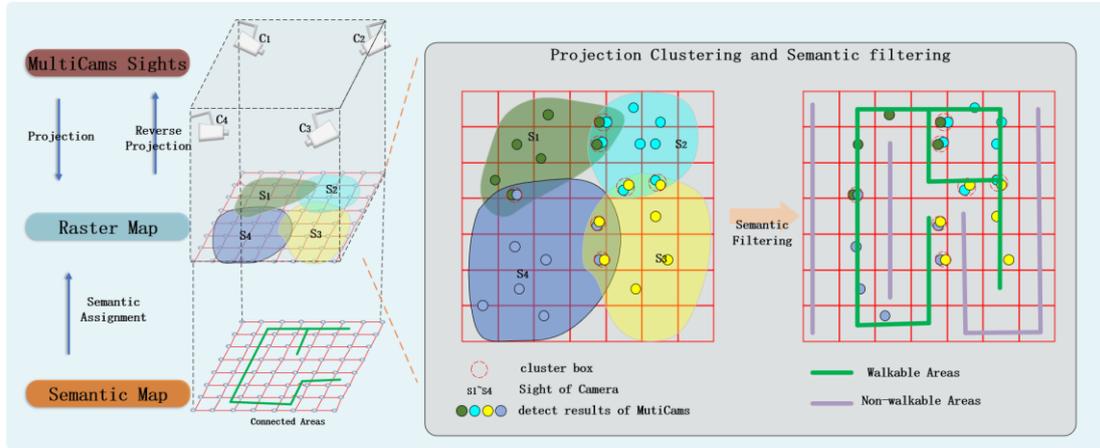

**Figure 6**  Co-visibility semantic information coarse matching

## (2) Walkability-based probabilistic raster normalization algorithm

Using the position encoding input, a 3×3 rectangular region centered on the target location is selected, and a probabilistic model with Gaussian kernel function is taken to construct the target probabilistic raster. Based on the feasibility of the raster semantic map, the probability raster is constrained to set the probability of the overlapping region to 0 when the probability raster intersects with the infeasible region while normalizing the overall probability raster, as shown in Figure 7.

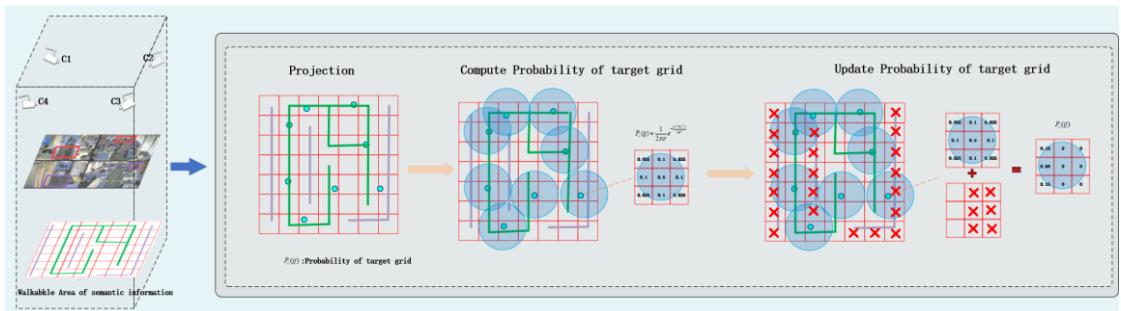

**Figure 7**  Walkability semantic information

## (3) Calculation of target transfer probabilities based on connectivity combined with Markov random fields

Based on the location-encoded input, the global Markov random field [15] is constructed with the camera as the node to calculate the transfer probability of the current target in the raster semantic map, as shown in Figure 8. An n-step transfer matrix $P(N)$ can be obtained, and the probability of a target located at camera i reaching camera j is $P_{ij}$.

$$P(N) = \begin{bmatrix} P_{11}(N) & P_{12}(N) & \cdots & P_{1n}(N) \\ P_{21}(N) & P_{22}(N) & \cdots & P_{2n}(N) \\ \cdots & \cdots & \cdots & \cdots \\ P_{n1}(N) & P_{n2}(N) & \cdots & P_{nn}(N) \end{bmatrix} \quad (1)$$

$$P_{ij} = argmaxP_{ij}(N) \quad (2)$$

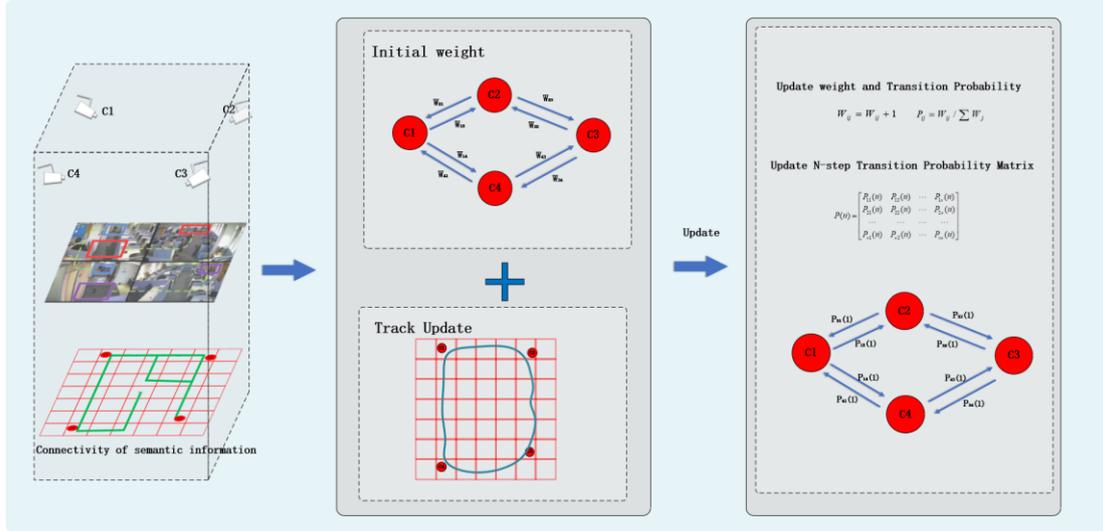

**Figure 8** Semantic information on connectivity

**2.2.3.2 Dynamic multi-dimensional feature matching algorithm combined with raster semantic map filtering**

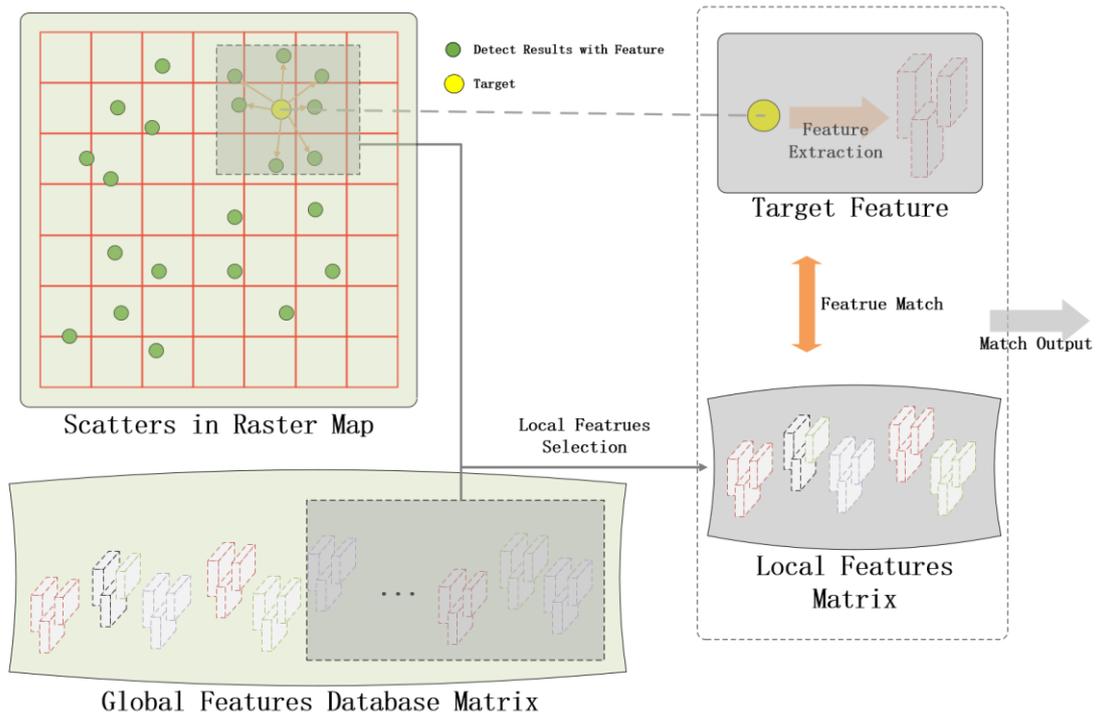

**Figure 9** Dynamic feature matching

As shown in Figure 9, based on the position encoding input, a subset of the global feature library is obtained using the target location relationship in the local area of the raster semantic map, and a local features matrix is constructed in real time. Using the multi-dimensional feature vector input, the local features matrix is combined with the local features matrix for similarity matching to obtain the target best matching ID results.

**2.2.3.3 Space-time logic-based multi-visual target tracking algorithm**

**(1) Single camera field of view based target tracking mechanism**

In this study, based on the continuous frame position encoding input, the bipartite graph optimal matching operator [30] is used as the core to achieve complete matching of front and back frames and obtain the target tracking results of continuous detection frames in a single camera field of view to achieve uniformity of ID and confidence.

**(2) Detection frames and confidence transfer based on a cross-camera overlapping field of view scenes**

A weighted interjection mechanism was developed to calculate the Euclidean distance between multi-target detection results between cameras based on cross-camera position coding input and set a dynamic threshold upper limit. When the Euclidean distance between the two inter-camera results is the smallest and the distance is less than the upper dynamic threshold, the two targets are considered aligned. The aligned detection results will share the same ID and confidence score, which means the one with higher confidence score will overwrite the other one. Hence, alignment of the multi-camera view tracking trajectory is achieved.

**(3) Re-recognition judgment based on a cross-camera non-overlapping field of view scenes**

To construct a global probability transfer matrix, a topology diagram was built using the space-time correlation between cameras, defining the nodes as camera fields of view and the edges as transfer probabilities. When the target disappears from the camera field of view, the global probability transfer matrix can be used to predict which field of view the target will reappear. For instance, the following figure consisted of four nodes with transfer probability labeled on edges.

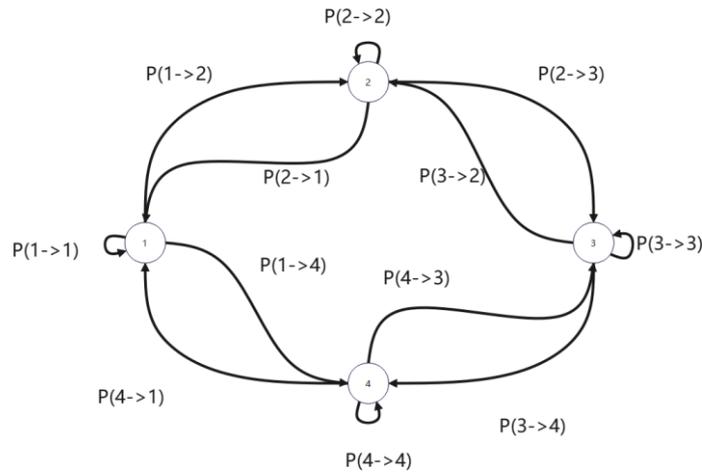

**Figure 10**      Camera probability transfer topology

St-Reid was applied to extract the tracklet image information and space-time information. Given tracklets $L_i, L_j$ and their feature vectors $x_i, x_j$, $P_a$ denotes the probability of similarity of image information between $L_i$ and $L_j$. $P_t$ denotes the probability of space-time association between $L_i$ and $L_j$ that is, whether the target in the last frame of $L_i$ and the target in the first frame of $L_j$ belongs to the same object. So the probability density function of $P_t$ is related to camera ID, target ID and time. $P_a$ and $P_t$ are given by the equations:

$$P_a(L_i, L_j) = \frac{x_i x_j}{|x_i||x_j|} \quad (3)$$

$$P_t(L_i, L_j) = P(m_{i1} = m_{i2}|c_{j1}, c_{j2}, t_{k2} - t_{k1}) \quad (4)$$

In which $m_{i1}$, $m_{i2}$ are the target ID, $c_{j1}$, $c_{j2}$ are the camera ID, $t_{k2}$ is the time the target $m_{i1}$ leaves the camera $c_{j1}$, and $t_{k2}$ is the time the target $m_{i2}$ appears in the camera $c_{j2}$.

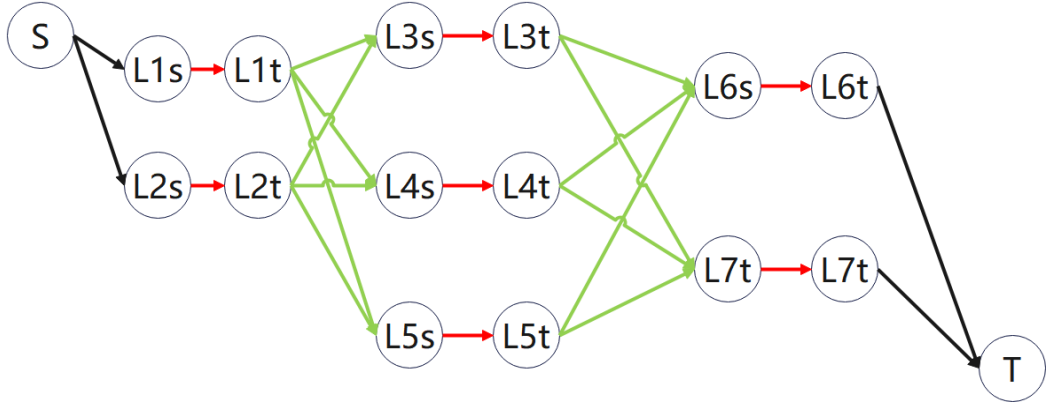

**Figure 11**  Global graph model (EGM), 3 steps, 7 tracklets

The Markov random field [31] is used as the base model to construct the global graph with the start and end of the tracklet as nodes. The definitions and weights of the edges of the global graph are as follows.

a) The red edges indicate the connection between the start moment $t_i^e$ and the end moment $t_i^s$ of the tracklet. The probabilities and weights are computed by the following expressions:

$$c_i = P(L_i|\Gamma) = \frac{\sum_{k=t_i^s}^{t_i^e} \alpha_k}{t_i^e - t_i^s} \quad (5)$$

$$w(L_i) = -\log \frac{c_i}{1-c_i} \quad (6)$$

where $L_i$ denotes the tracklet, $\Gamma$ is the set of trajectories, $\alpha_k$ is the probability of similarity between the frames of the tracklet, $c_i$ is the probability that the tracklet $L_i$ holds, and $w(L_i)$ is the probability that the tracklet $L_i$ is the weight of the tracklet in the graph.

b) The green edges indicate connections between tracklets given by the following weights.

$$w(L_i|L_j) = -k_a \log P_a(L_iL_j) - k_t \log P_t(L_iL_j) \tag{7}$$

c) The black edges indicate edges connected at the start and end nodes with a weight of zero.

The min-cost flow method[42] is used to obtain the relationships between tracklets and the corresponding IDs. The optimal set of tracks can be calculated using the following:

$$\begin{aligned} \Gamma^* &= \arg\max\nolimits_\Gamma \prod_i P(L_i|\Gamma) \prod_{\Gamma_k \in \Gamma} P(\Gamma_k) \\ &= \arg\max\nolimits_\Gamma \prod_i P(L_i|\Gamma) \prod_{\Gamma_k \in \Gamma} \prod_{L_{k_1}, L_{k_2}, \ldots \in \Gamma_k} P(L_{k_{j+1}}|L_{k_j}) \\ &\quad \Gamma_i \cap \Gamma_j = \Phi, \forall\, i \neq j \end{aligned} \tag{8}$$

Using this approach, the tracking association for multiple tracklets in a cross-camera non-overlapping field-of-view scenario can be determined, enabling the transfer of IDs.

**2.2.3.4 Construction of a Space-time Convergence Network**

The ReID problem is resolved by iterative delivery of tracking ensemble queries, while the generation and delivery of tracking ensemble queries require consideration of space-time correlation, image continuity, and other issues. In this study, a space-time convergence network is constructed with enhanced temporal correlation to provide contextual a priori information for continuous target tracking, as shown in Figure 12 below.

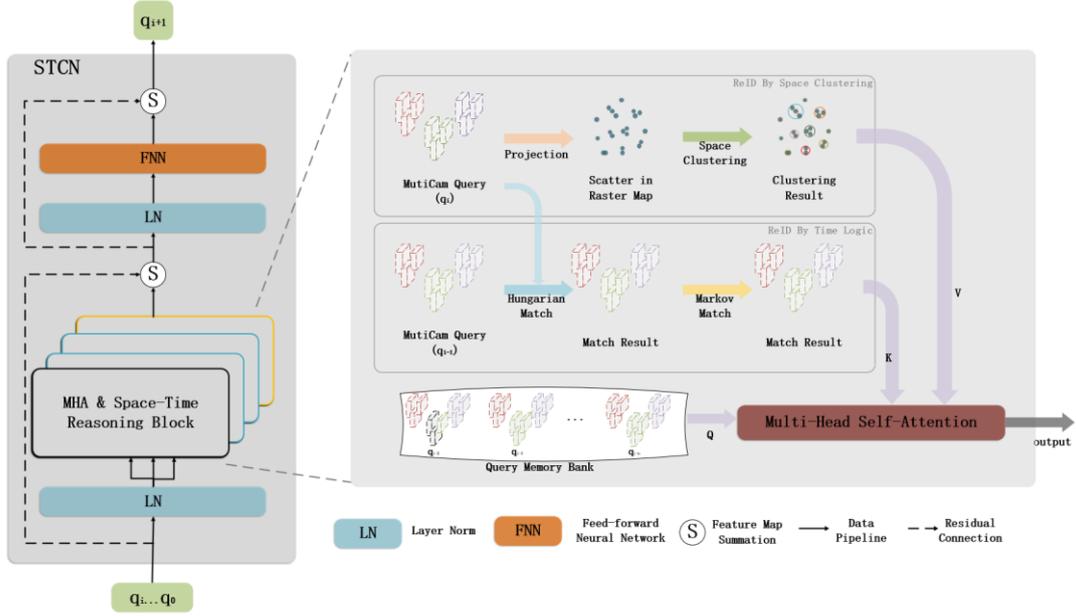

**Figure 12**　　STCN structure diagram

The STCN involves the following steps：

a) Spatial clustering based on a multi-camera query was used to obtain query input under spatial logic.

b) Hungarian matching algorithm and Markov random field matching based on the query of consecutive frames were incorporated to obtain the query input under temporal logic.

c) Based on inputs 1) and 2), a query ensemble $q_{bank}$ is constructed in combination with a historical continuous frame tracking query, while a cascade of multiple queries is implemented as follows:

$$q_{bank} = \{\tilde{q}_c^{i-M}, \dots, \tilde{q}_c^i\}$$
$$tgt = \tilde{q}_c^{i-M} \oplus \cdots \tilde{q}_c^{i-1} \oplus \tilde{q}_c^i \tag{9}$$

where $tgt$ denotes a cascade of queries.

d) The cascaded queries are fed into the multi-attention module to generate attention weights, resulting in the following dot product attention formula.

$$q_{sa}^i = \sigma_s\left(\frac{tgt \cdot tgt^T}{\sqrt{d}}\right) \cdot \bar{q}_c^i \tag{10}$$

where $\bar{q}_c^i$ is taken as the multiple attention (MHA) query, $\sigma_s$ denotes the softmax function, and $d$ indicates the dimension of the track query.

e) Further tuning and optimization based on the feed-forward network (FFNN) is employed to finally output the track query $q_t^{i+1}$ for the next frame.

$$\begin{aligned} t\tilde{g}t &= LN(q_{sa}^i + \bar{q}_c^i) \\ \hat{q}_c^i &= LN(FC(\sigma_r(FC(\widetilde{g}\tilde{g}t))) + t\tilde{g}t) \end{aligned} \tag{11}$$

where $FC$ denotes a linear projection layer, and $LN$ denotes the layer normalization.

**2.2.4 Construction of a retrospective mechanism based on inverse order processing**

In this study, there are multiple dimensions of the target re-identification mechanism. In the positive-order real-time processing, there is a situation where the continuous tracking target IDs are only optimally matched in the middle section of the tracklet, in which case, there are unmatched IDs in the historical tracklet, thus affecting the overall accuracy.

Therefore, a retrospective mechanism is constructed (see Figure 13) to process the relevant tracklet in reverse order after completing the overall tracking and optimize the historical target IDs.

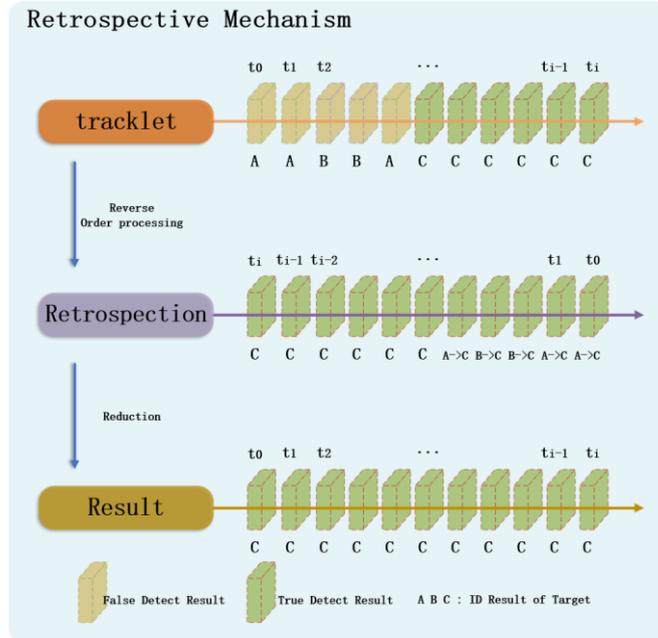

**Figure 13**　　Schematic diagram of tracklet-based retrospective processing

### 2.2.5 Collective Average Loss

In this study, we constructed a self-attentive model based on STCN for spatial-temporal modelling of multi-target tracking query, and performed loss calculation based on continuous multi-frame tracking prediction results with the following loss function [10].

$$CAL = \frac{\sum L_i}{\sum V_i}$$

$$L_i = \lambda_{cls} L_{cls} + \lambda_{L_1} L_{L_1} + \lambda_{iou} L_{iou} = \lambda_{cls} (\log(P_{feature} * P_{st} * P_{map})) + \lambda_{L_1} \| b_i - \acute{b}_i \| + \lambda_{iou} C_{iou}(b_i, \acute{b}_i) \quad (12)$$

where CAL denotes collective average loss, $V_i$ is the total number of targets, $\lambda_{cls}$, $\lambda_{L_1}$, $\lambda_{iou}$ are the weight parameters, $P_{feature}$, $P_{st}$, $P_{map}$ are the feature matching probability, spatio-temporal matching probability and semantic map matching probability respectively, $b_i$ is the current detection frame, $\acute{b}_i$ is the real detection frame, and $C_{iou}$ is the result of the IoU loss function calculation.

## 2.3 Evaluation Metrics

The Identification F1-score (IDF1)[32-33], the Multiple Object Tracking Accuracy (MOTA)[34-37], and the Multi-camera Tracking Accuracy (MCTA) [43-47]were used as the outcome criteria in this study.

a) IDF1, the summed average of Identification Precision (IDP) and Identification Recall Rate (IDR), was used to assess the degree of identification accuracy.

b) MOTA was used to assess the reliability of target tracking.

c) MCTA, a multi-camera tracking metric, was used to assess the tracking accuracy of multiple cameras.

## 3  Experiments

## 3.1 Implementation details

The above method is implemented and evaluated on MOT17 and OVIT-MOT01. MOT17 is a public dataset used for accuracy validation and cross-sectional comparison with other algorithms. OVIT-MOT01 is a self-built dataset for validating single and multi-camera target tracking accuracy, comparing the annotation results, and generating the accuracy reports. Moreover, an ablation study is presented to verify the contribution of each components.

## 3.2 Validation of single camera accuracy results based on the publicly available dataset (MOT17)

Accuracy validation was performed using the MOT17 dataset, and the results are summarized in Table 1. According to Table 2, the IDF1 value of this research method reaches 0.782 and the MOTA value 0.836 in the public dataset MOT17. Comparing with the public algorithms in the current MOT17 ranking, the MOTA accuracy of this study is ranked first and the IDF1 value is ranked fourth, which shows that the target tracking algorithm of this research can reach the average level of the current leading algorithms.

**Table 1-** MOTA values based on MOT17

|  | IDF1 | MOTA | IDP | IDR | Recall | Precision |
|---|---|---|---|---|---|---|
| **MOT17-02-SDP** | 0.577433 | 0.666111 | 0.682192 | 0.500565 | 0.707013 | 0.963547 |
| **MOT17-04-SDP** | 0.907841 | 0.945097 | 0.919049 | 0.896903 | 0.961099 | 0.984831 |
| **MOT17-05-SDP** | 0.735971 | 0.788926 | 0.809045 | 0.675004 | 0.816684 | 0.97886 |
| **MOT17-09-SDP** | 0.643427 | 0.782535 | 0.696413 | 0.597934 | 0.824601 | 0.960411 |

| | | | | | |
|---|---|---|---|---|---|
| MOT17-10-SDP | 0.648384 | 0.730119 | 0.716602 | 0.592024 | 0.784952 | 0.950127 |
| MOT17-11-SDP | 0.835397 | 0.873145 | 0.860902 | 0.811361 | 0.910237 | 0.965816 |
| MOT17-13-SDP | 0.727051 | 0.801151 | 0.767242 | 0.690861 | 0.855437 | 0.950014 |
| **OVERALL** | 0.781575 | 0.83606 | 0.828008 | 0.740073 | 0.868322 | 0.971496 |

The results of the proposed method are compared with other published algorithms, and the comparative summary is presented in Table 2.

Table 2- Comparison of open algorithm metrics based on MOT17

| Method | IDF1 | MOTA |
|---|---|---|
| SelfAT[32] | 0.798 | 0.800 |
| ByteTrack[33] | 0.773 | 0.803 |
| QuoVadis[34] | 0.777 | 0.803 |
| FOR_Tracking[35] | 0.777 | 0.804 |
| BoT_SORT[36] | **0.802** | 0.805 |
| BYTEv2[37] | 0.789 | 0.806 |
| MiniTrackV2[49] | 0.788 | 0.818 |
| Ours | 0.782 | **0.836** |

## 3.3 Continuous tracking accuracy based on the self-built dataset OVIT-MOT01

Table 3 shows the continuous tracking accuracy results in single-camera scenes using the OVIT-MOT01 dataset.

Table 3- OVIT-MOT01 MOTA values for each camera

| Transformer | Feature | Logic | Cam1 | Cam2 | Cam3 | Cam4 | Cam5 | Total |
|---|---|---|---|---|---|---|---|---|
| √ | | | 0.898 | 0.757 | 0.904 | 0.729 | 0.634 | 0.819 |
| √ | √ | | 0.898 | 0.757 | 0.905 | 0.729 | 0.772 | 0.853 |
| √ | √ | √ | 0.910 | 0.772 | 0.914 | 0.743 | 0.760 | 0.860 |

Note: The MOTA values for each of the five cameras were tested in three cases (a. transformer temporal logic tracking; b. transformer + multi-dimensional feature dynamic matching; c. transformer + multi-dimensional feature dynamic matching + temporal logic matching).

### 3.4 Ablation experiments based on OVIT-MOT01

The overall cross-camera re-recognition ablation experiments of the five cameras was tested in four scenarios (a. transformer temporal logic tracking; b. transformer + multi-dimensional feature dynamic matching; c. transformer + multi-dimensional feature dynamic matching + temporal logic matching; d. transformer + multi-dimensional feature dynamic matching + temporal logic matching + retrospective mechanism); the summary of results is presented in Table 4.

Table 4- Results of ReID ablation experiments based on a self-constructed dataset

| Transform | Feature | Logic | Back-view | IDF1 | MOTA | MCTA |
|---|---|---|---|---|---|---|
| √ | | | | 0.417 | 0.819 | / |
| √ | √ | | | 0.933 | 0.853 | 0.860 |
| √ | √ | √ | | 0.948 | 0.860 | 0.878 |
| √ | √ | √ | √ | 0.981 | 0.863 | 0.909 |

## 4 Discussion

The MOTA for the single-camera target tracking using the MOT17 dataset reached 0.782, while the IDF1 value was 0.836 (see Tables 1 and 2). Since the focus of this study is on complex scenarios with multiple cameras and scales, the subsequent analyses were based on the OVIT-MOT01 dataset.

### 4.1 Transformer-based temporal logic tracking matching

Using the transformer's temporal logic tracking matching, the overall MOTA for the OVIT-MOT01 dataset reached 0.819 while the IDF1 value was only 0.417 (see Tables 3 and 4). The image accuracy results were affected by the following factors: (1) ID switching due to masking; (2) ID switching due to the target entering and leaving the camera; (3) target ID switching due to perspective and scale shifts, as detailed in Figure 14(a), Figure 14(b) and Figure 14(c).

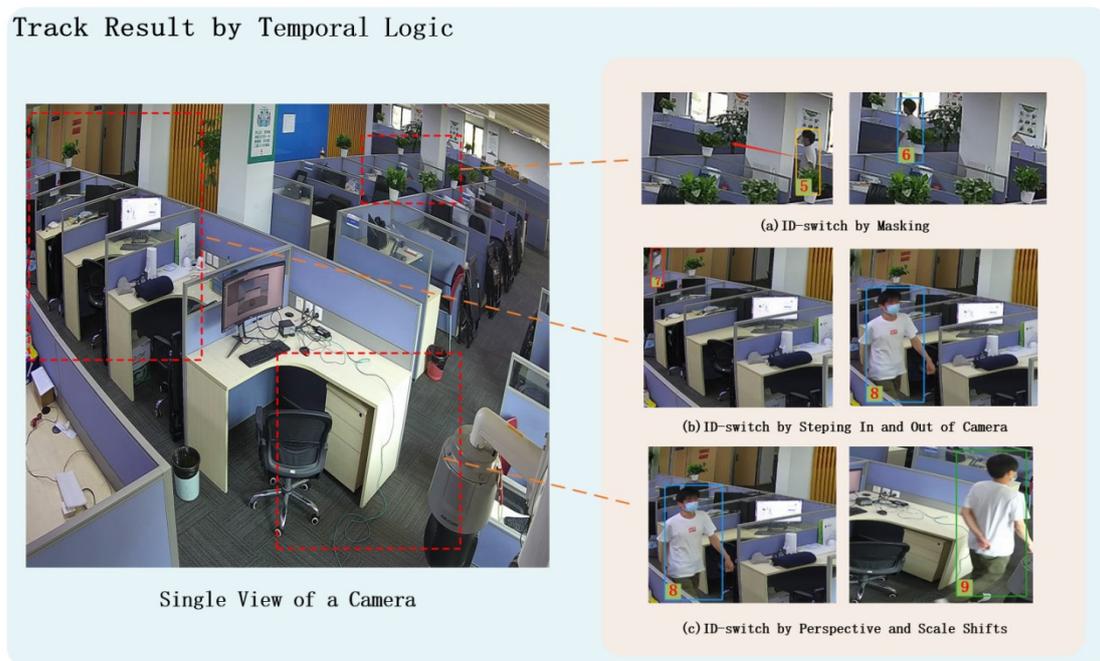

**Figure 14**　　Graph of continuous tracking results based on transformer's temporal logic

## 4.2 Transformer adds multi-dimensional feature matching

With the addition of multi-dimensional dynamic feature matching, the overall MOTA using the OVIT-MOT01 dataset improved to 0.853, the MCTA reached 0.860, and the IDF1 significantly increased to 0.933 (see Table 4). The feature-based matching is independent of time and space and can provide accurate re-identification of targets. The constructed multi-dimensional feature library effectively compensates for the problem of false detection caused by changes in illumination, angle, and scale. After matching, targets lost due to occlusion can be re-tracked after re-emergence. The ID switch situation generated by targets when crossing cameras is significantly reduced, and the ID swap due to target interleaving is eliminated, resolving the target loss problem caused by viewpoint changes. As shown in Figure 15, the ID switches due to masking, cross-camera, perspective shifts, and scale shifts in Figure 14 have been optimized.

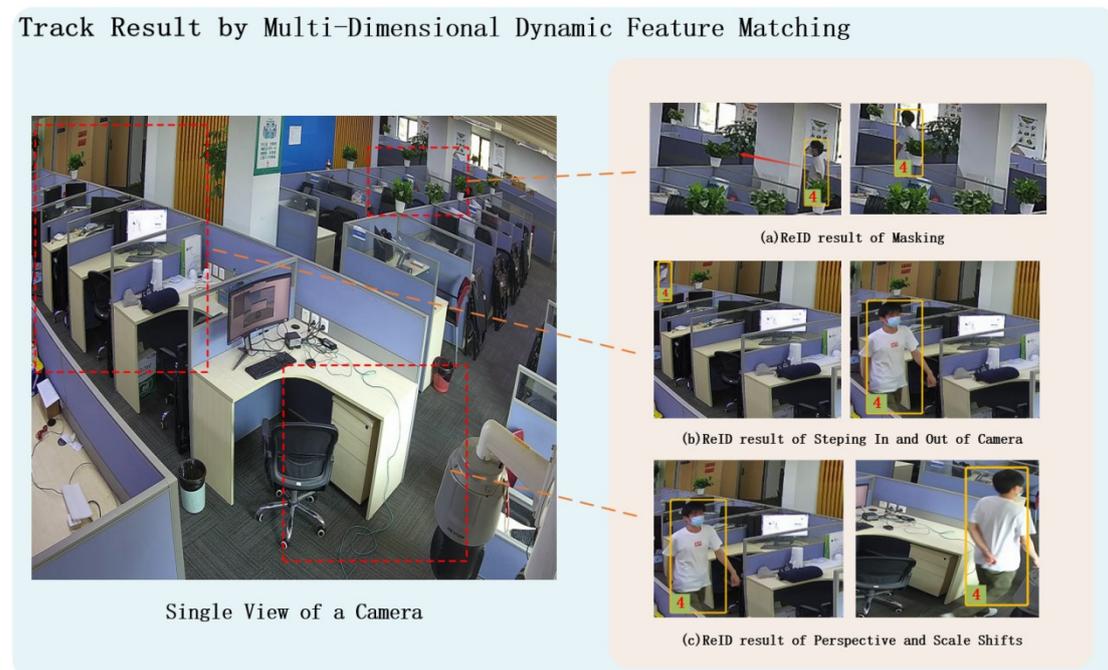

**Figure 15**    Graph of results after adding multi-dimensional dynamic feature matching

## 4.3 Transformer adds temporal logic matching

With the addition of space-time logic matching, the overall MOTA improved to 0.860, while the IDF1 increased to 0.948 and MCTA to 0.878, as detailed in Table 4. Under the global raster semantic map, with spatial logic matching, the detection accuracy of the overlapping field of view area is improved. The spatial position association between the body and head detection frames is introduced so that when the body detection frame gets lost due to the target's intersection, occlusion, or scale change, the head detection can still maintain the continuous tracking and ID of the target.

As shown in Fig. 16(a), before the addition of logical matching, the interleaved occlusion result for target ID_2 was mislabeled as ID_3. But after adding space-time logical matching, the tracking ID for target ID 2 was kept unchanged even though it was interleaved due to the continuous tracking based on head detection. As shown in Figure 16(b), the detection target features at the edges were not significant, causing ID recognition errors, while the addition of logical matching resulted in the correct target matching across cameras due to spatial clustering.

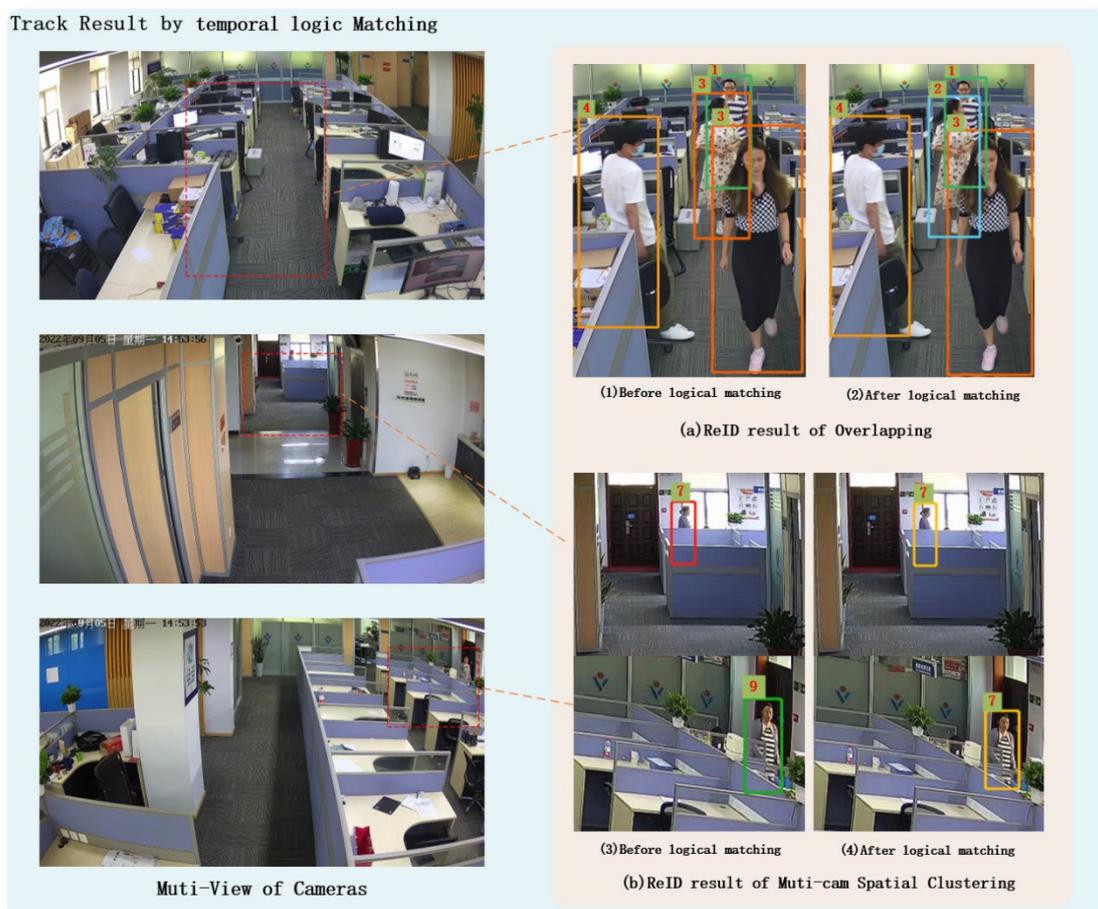

**Figure 16**  Graph of results after adding temporal logic matching

## 4.4 Transformer adds a retrospective mechanism

The addition of the retrospective mechanism optimizes the overall accuracy by using the continuous image sequence information in verifying and correcting some wrong and missed detections in the historical data. The experiment results show that the proposed method improved the overall MOTA to 0.863, IDF1 to 0.981 and MCTA to 0.909.

As shown in Figure 17, Figure 17(a) shows that the head-based ReID detection result was successfully obtained after retrospective mechanism. Figure 17(b) shows the results of sequential image processing tracking of consecutive frames before the retrospective mechanism, which shows that no ReID result was obtained for the head-based detection of target ID_2 at moments $t_0$ and $t_1$; Figure 17(c) shows the results of consecutive frames after the retrospective processing, which shows that the ReID was completed for target ID_2 at moments $t_0$ and $t_1$.

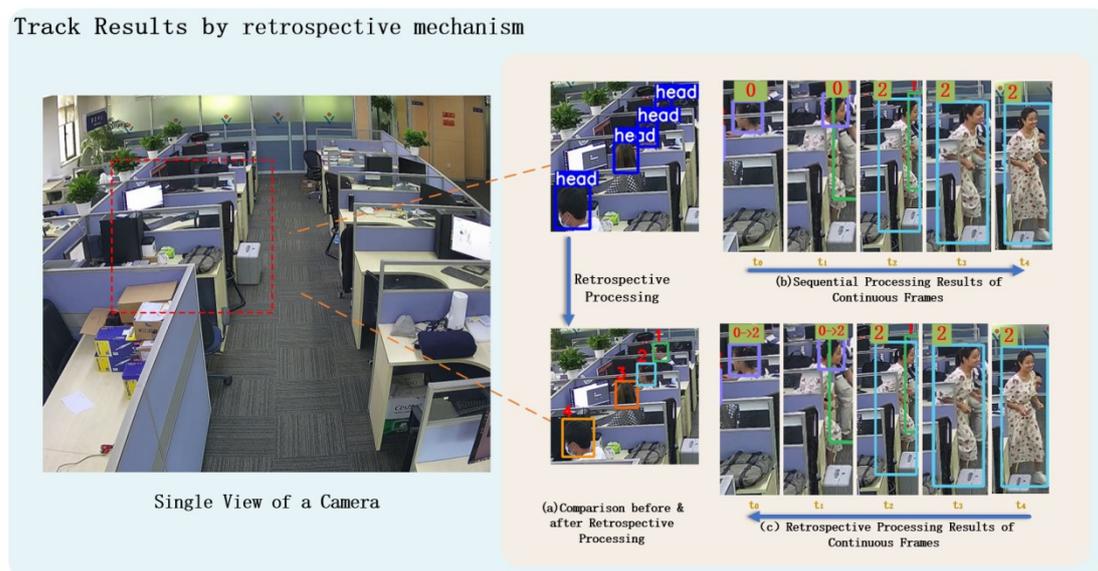

**Figure 17**  Optimization of head tracking results with the addition of the retrospective mechanism

## 5  Conclusion

An end-to-end multi-target tracking approach was developed based on the transformer's self-attention improvements. In the decoder, the cross-camera target re-recognition results for temporal logic, spatial logic, and feature matching are fused. A space-time convergence network (STCN) is constructed to pass the re-recognition parameters, while a backward-order processing mechanism is used to optimize the overall accuracy. The proposed approach achieves method normalization for complex multi-view and multi-scale scenarios and solves the multi-target tracking problem for overlapping and non-overlapping field-of-view scenarios. In the experiments using public datasets, the proposed approach performed well against other algorithms for single-camera tracking. For the MOT01 dataset, the approach method achieved a MOTA value of 0.863 and an IDF1 value of 0.981, significantly improving the overall accuracy and tracking effect.

The construction process for the raster semantic map in this method remains tedious. In the future, intelligent VR terminals can be combined with related algorithms, such as neural radiation field[38](NeRF), to achieve more rapid construction of semantic raster map. Other localization sources (e.g., audio[39], UWB[40], Bluetooth[41]) can also be introduced to assist in target tracking to improve reliability and robustness.

## Acknowledgments

The authors are sincerely grateful to the editors as well as the anonymous reviewers for their valuable suggestions and comments that helped us improve this paper significantly.

## Data availability statement

The data that support the findings of this study are available from the corresponding author upon reasonable request.

## Disclosure statement

No potential conflict of interest was reported by the author(s).

## Funding

This work was supported by the Key Research & Development of Hubei Province (2020BIB006);The Natural Science Foundation of Hubei Province(2020CFA001);The Key Research & Development of Hubei Province (2020AAA004).

## Notes on contributors

**Yong Hong** is a doctoral student of LIESMARS,Wuhan University.His research direction is multi-source fusion perception and navigation positioning.

**Deren Li** received the Ph.D. degree in photogrammetry from the University of Stuttgart, Germany, in 1986. He is a scientist in surveying, mapping and remote sensing from Wuhan University, China. He enjoys dual memberships of both Chinese Academy of Sciences and Chinese Academy of Engineering. He is also the member of International Eurasia Academy of Sciences and International Academy of Astronautics. He received doctor degree from University of Stuttgart and honorary doctorate from ETH Zürich. International Society for Photogrammetry and Remote Sensing awarded him the Honorary Member and the Brock Gold Medal in recognition of outstanding contributions to photogrammetry. So far, his publications include 11 monographs, over 700 journal articles and over 300 colloquium papers, which have been cited for more than 27000 times.

**Shupei Luo** graduated from Huazhong University of Science and Technology in 2019, majoring in Electronic Information Engineering. His research interests include multi-target multi-camera tracking and vehicle-road collaboration.

**Xin Chen** graduated from Wuhan Institute of Technology with a major in Computer science. His research interests include computer graphics, slam, and target tracking.

**Yi Yang** received the Master of Science degree in Electrical and Computer Engineering from the University of Illinois at Urbana-Champaign. His research interests include vision transformer and neural graphics.

**Mi Wang**, PhD, professor of LIESMARS, Wuhan University, is mainly engaged in image processing for digital photogrammetry and remote sensing of new-type airborne and spaceborne imaging sensors.

## ORCID

Yong Hong https://orcid.org/0000-0003-1290-849X